\documentclass[11pt]{article}

\usepackage[utf8]{inputenc} 
\usepackage[T1]{fontenc}    
\usepackage{hyperref}       
\usepackage{url}            
\usepackage{booktabs}       
\usepackage{amsfonts}       
\usepackage{textcomp}

\usepackage{graphicx}

\usepackage{algorithm}
\usepackage{algpseudocode}

\usepackage{amsmath}
\usepackage{bm}
\usepackage{amssymb}
\usepackage{amsthm}

\usepackage[a4paper, left=3cm, right=2.5cm, top=3cm, bottom=3cm]{geometry}

\usepackage{authblk}
\usepackage{natbib}
\usepackage{doi}

\newtheorem{lma}{Lemma}
\newtheorem{thm}{Theorem}

\newcommand{\R}{\mathbb{R}}
\newcommand{\N}{\mathbb{N}}
\newcommand{\enc}{f}
\newcommand{\dec}{g}
\newcommand{\Enc}{\phi}
\newcommand{\Dec}{\psi}
\newcommand{\out}{h}
\newcommand{\tre}[1]{\hat{#1}}

\begin{document}

\title{Recursive Tree Grammar Autoencoders}

\author[1]{Benjamin Paaßen}
\author[1]{Irena Koprinska}
\author[1]{Kalina Yacef}
\affil[1]{School of Computer Science\\The University of Sydney}

\date{This is a preprint of a paper submitted to the ECML-PKDD2022 Journal Track.} 
\pagestyle{myheadings}
\markright{Preprint of a paper submitted to the ECML-PKDD2022 Journal Track.}

\maketitle

\begin{abstract}
Machine learning on trees has been mostly focused on trees as input to algorithms.
Much less research has investigated trees as output, which has many applications, such as 
molecule optimization for drug discovery, or hint generation for intelligent tutoring systems.
In this work, we propose a novel autoencoder approach, called recursive tree grammar 
autoencoder (RTG-AE), which encodes trees via a bottom-up parser and decodes trees via
a tree grammar, both learned via recursive neural networks that minimize the variational 
autoencoder loss.
The resulting encoder and decoder can then be utilized in subsequent tasks,
such as optimization and time series prediction.
RTG-AEs are the first model to combine variational autoencoders, grammatical knowledge, and recursive processing.
Our key message is that this unique combination of all three elements
outperforms models which combine any two of the three.
In particular, we perform an ablation study to show that our proposed method improves the
autoencoding error, training time, and optimization score on synthetic
as well as real datasets compared to four baselines.
\end{abstract}


\section{Introduction}

Deep neural networks on trees have made significant progress in recent years
with novel models that achieved unprecedented performance on tree-related tasks,
such as tree echo state networks \citep{Gallicchio2013}, tree LSTMs \citep{Tai2015},
code2vec \citep{Alon2019}, or models from the graph neural network family \citep{Kipf2017,Micheli2009,Scarselli2009}.
However, these advancements are mostly limited to tasks with \emph{numeric output}, such as
classification and regression. By contrast, much less research has focused on tasks that require
\emph{trees as output}, such as as molecular design \citep{Kusner2017,Jin2018}
or hint generation in intelligent tutoring systems \citep{Paassen2018JEDM}. In this work,
we propose a novel autoencoding model which can encode a tree $\tre x$
into a vector encoding $\vec z$ and decode a vector $\vec z$ back into 
a tree $\tre x$. Thus, our model supports not only classification and regression
tasks but also optimization over trees or time series prediction on trees.

\begin{figure}
\begin{center}
\includegraphics[width=8cm]{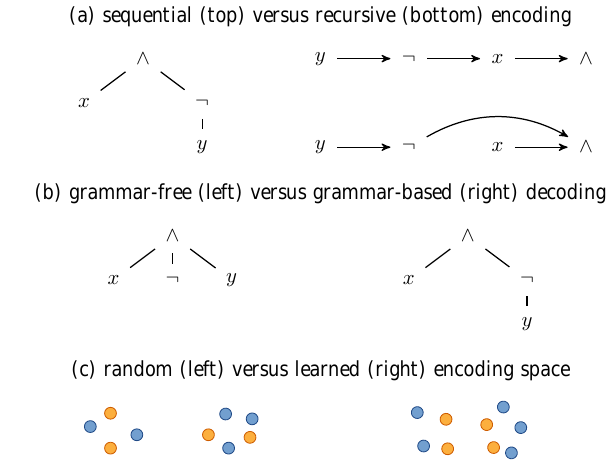}
\end{center}
\caption{An illustration of the advantages (a) of recursive over sequential processing, (b) of utilizing grammatical knowledge, and (c) of learning the
encoding end-to-end. In c), each point represents the encoding of a tree
and color indicates some semantic attribute with respect to which the encoding space should be smooth (right).}
\label{fig:processing}
\end{figure}

There are three key roots for our approach in prior literature.
First, \citet{Kusner2017} represented strings as a sequence of context-free grammar rules, encoded this rule sequence in the latent space of a variational autoencoder using a convolutional neural
net, and decoded back into a rule sequence with a gated recurrent unit (GRU) \citep{Cho2014}.
We believe that, for trees, the sequential processing proposed by \citet{Kusner2017}
is insufficient because it introduces long-range dependencies. For example,
when processing the tree $\wedge(x, \neg(y))$, a sequential representation
would be $y, \neg, x, \wedge$. When processing $\wedge$, we want to take
the information from its children $x$ and $\neg$ into account, but $\neg$
is already two steps away. If we replace $x$ with a large subtree, this
distance can become arbitrarily large. To avoid such long-range dependencies,
we propose recursive processing, where the information
flow follows the structure of the tree (Figure~\ref{fig:processing}, a) \citep{Tai2015,Pollack1990,Sperduti1997}. Therefore, we call our approach recursive tree grammar autoencoders (RTG-AEs).

Second, \citet{Zhang2019} suggested a variational autoencoder for graphs by representing
a graph as a sequence of node and edge insertions, which in turn can be encoded and decoded with a recurrent
neural network. While this approach is directly applicable to trees, it
needs to learn all grammatical structure of a domain, in particular the appropriate
number and order of children in each context. This can lead to obvious
errors, such as decoding the tree $\wedge(x, \neg(y))$ as $\wedge(x, \neg, y)$, which is clearly ungrammatical (Figure~\ref{fig:processing}, b).
Instead, in line with \citet{Kusner2017}, we believe that grammatical knowledge
is crucial as an inductive bias to prevent easily avoidable decoding mistakes.
In contrast to \citet{Kusner2017}, we employ tree grammars instead of string grammars.

Finally, in our own previous work \citep{Paassen2020IJCNN}, we implemented a recursive tree grammar neural network architecture
using tree echo state networks \citep{Gallicchio2013}. This shallow learning
approach utilizes a random, high-dimensional and scattered encoding space.
In this work, we propose to learn the entire autoencoder end-to-end via
a variational autoencoding approach, thus yielding a lower-dimensional,
smoother encoding space in which sampling and optimization is easier (Figure~\ref{fig:processing}, c).

As such, RTG-AEs combine variational autoencoders, grammatical knowledge, and recursive 
processing. Our key message is that this combination outperforms approaches
which combine only two of these three aspects.
In particular, we compare to the model of \citet{Kusner2017} and to
an ablated version of RTG-AE which combine variational autoencoding with grammatical knowledge but process the trees sequentially;
we compare to the DVAE of \citet{Zhang2019} which combines variational 
autoencoding with recursive processing but does not use grammatical knowledge; and we compare to TES-AE 
\citep{Paassen2020IJCNN}, which is an ablated version of RTG-AE
that combines grammatical knowledge with recursive processing but only trains
the decoding layers, while we propose end-to-end learning.
We evaluate RTG-AEs and all baselines on four data sets, including
two synthetic and two real-world ones, including hyperparameter optimization and crossvalidation to
give more robust performance estimates compared to prior work. We also run optimization with multiple 
repeats to provide more robust results.
Additionally, our paper contributes a correctness proof of our encoding based on the theory of
regular tree grammars.

We begin by discussing background and related work before we introduce the RTG-AE architecture
and evaluate it on four datasets, including two synthetic and two real-world ones.

\section{Background and Related Work}

Our contribution relies on substantial prior work, both from theoretical computer science and
machine learning. We begin by introducing our formal notion of trees and tree grammars, after
which we continue with neural networks for tree representations.

\subsection{Regular Tree Grammars}
\label{sec:rtg}

Let $\Sigma$ be some finite alphabet of symbols. We recursively define a \emph{tree} $\tre x$ over
$\Sigma$ as an expression of the form $\tre x = x(\tre y_1, \ldots, \tre y_k)$,
where $x \in \Sigma$ and where $\tre y_1, \ldots, \tre y_k$ is a list of trees over $\Sigma$.
We call $\tre x$ a \emph{leaf} if $k = 0$, otherwise we call $\tre y_1, \ldots, \tre y_k$
the \emph{children} of $\tre x$. We define the \emph{size} $\lvert \tre x\rvert $ of a tree $\tre x = x(\tre y_1, \ldots, \tre y_k)$ as $1 + \vert \tre y_1\vert  + \ldots + \vert \tre y_k\vert $.

Next, we define a \emph{regular tree grammar} (RTG) \citep{Brainerd1969,Comon2008}
$\mathcal{G}$ as a $4$-tuple $\mathcal{G} = (\Phi, \Sigma, R, S)$,
where $\Phi$ is a finite set of nonterminal symbols, $\Sigma$ is a finite alphabet as before,
$S \in \Phi$ is a special nonterminal symbol which we call the starting symbol, and $R$ is a
finite set of production rules of the form $A \to x(B_1, \ldots, B_k)$ where
$A, B_1, \ldots, B_k \in \Phi$, $k \in \N_0$, and $x \in \Sigma$.
We say a sequence of rules $r_1, \ldots, r_T \in R$ \emph{generates} a tree $\tre x$ from some
nonterminal $A \in \Phi$ if applying all rules to $A$ yields $\tre x$, as specified in Algorithm~\ref{alg:generate_tree}.
We define the regular tree language $\mathcal{L}(\mathcal{G})$ of grammar $\mathcal{G}$ as the set
of all trees that can are generated from $S$ via some (finite) rule sequence over $R$.

\begin{algorithm}
\caption{An algorithm for generating a tree $\tre x$ from a nonterminal $A$
and a sequence of production rules $r_1, \ldots, r_T$.}
\label{alg:generate_tree}
\begin{algorithmic}[1]
\Function{generate\_tree}{nonterminal $A$, rules $r_1, \ldots, r_T$}
\State Initialize a stack $S$ with $A$ on top.
\State Initialize a tree $\tre x \gets A()$.
\For{$t \gets 1, \ldots, T$}
	\State Pop nonterminal $A'$ from $S$.
	\State Let $r_t = A \to x(B_1, \ldots, B_k)$.
	\If{$A \neq A'$}
		\State Process fails.
	\EndIf
	\State Replace $A'$ in the tree with $x(B_1, \ldots, B_k)$.
	\State Push $B_k, \ldots, B_1$ onto $S$.
\EndFor
\If{$S$ is not empty}
	\State Process fails.
\EndIf
\State \Return $\tre x$.
\EndFunction
\end{algorithmic}
\end{algorithm}

The inverse of generation is called \emph{parsing}. In our case, we rely on the
bottom-up parsing approach of \citet{Comon2008}, as shown in Algorithm~\ref{alg:parse_tree}. For the input tree
$\tre x = x(\tre y_1, \ldots, \tre y_k)$, we first parse all children,
yielding a nonterminal $B_j$ and a rule sequence $\bar r_j$ that generates child $\tre y_j$ from $B_j$.
Then, we search the rule set $R$ for a rule of the form $r = A \to x(B_1, \ldots, B_k)$
for some nonterminal $A$, and finally return the nonterminal $A$ as well as the rule sequence
$r, \bar r_1, \ldots, \bar r_k$, where the commas denote concatenation.
If we don't find a matching rule, the process fails. Conversely, if the algorithm returns
successfully, this implies that the rule sequence $r, \bar r_1, \ldots, \bar r_k$
generates $\tre x$ from $A$. Accordingly, if $A = S$, then
$\tre x \in \mathcal{L}(\mathcal{G})$.

\begin{algorithm}
\caption{An algorithm for parsing a tree $\tre x$ according
to a regular tree grammar $\mathcal{G} = (\Phi, \Sigma, R, S)$.}
\label{alg:parse_tree}
\begin{algorithmic}[1]
\Function{parse\_tree}{tree $\tre x = x(\tre y_1, \ldots, \tre y_k)$}
\For{$j \gets 1, \ldots, k$}
	\State $B_j, \bar r_j \gets$ \Call{parse\_tree}{$\tre y_k$}.
\EndFor
\If{$\exists A \in \Phi : A \to x(B_1, \ldots, B_k) \in R$}
	\State $r \gets A \to x(B_1, \ldots, B_k)$.
	\State \Return $A$, $(r, \bar r_1, \ldots, \bar r_k)$.
\Else
	\State Process fails.
\EndIf
\EndFunction
\end{algorithmic}
\end{algorithm}

Algorithm~\ref{alg:parse_tree} can be ambiguous if multiple nonterminals $A$
exist such that $A \to x(B_1, \ldots, B_k) \in R$ in line 5.
To avoid such ambiguities, we impose that our regular tree grammars
are \emph{deterministic}, i.e.\ no two grammar rules have the
same right-hand-side. This is sufficient to ensure that any tree corresponds to
a unique rule sequence.

\begin{thm}
Let $\mathcal{G} = (\Phi, \Sigma, R, S)$ be a regular tree grammar.
Then, for any $\tre x \in \mathcal{L}(\mathcal{G})$ there exists exactly
one sequence of rules $r_1, \ldots, r_T \in R$ which generates $\tre x$.
\begin{proof}
Refer to Appendix~\ref{ap:unique}.
\end{proof}
\end{thm}

This is no restriction to expressiveness, as any regular 
tree grammar can be transformed into an equivalent, deterministic one.

\begin{thm}[Therorem 1.1.9 by \citet{Comon2008}]
Let $\mathcal{G} = (\Phi, \Sigma, R, S)$ be a regular tree grammar. Then, there exists a regular
tree grammar $\mathcal{G}' = (\Phi', \Sigma, R', \mathcal{S})$ with
a set of starting symbols $\mathcal{S}$ such that $\mathcal{G}'$ is deterministic and
$\mathcal{L}(\mathcal{G}) = \mathcal{L}(\mathcal{G}')$.

\begin{proof}
Refer to Appendix~\ref{ap:deterministic}.
\end{proof}
\end{thm}

It is often convenient to permit two further concepts in a
regular tree grammar, namely optional and starred nonterminals. In particular,
the notation $B?$ denotes a nonterminal with the production
rules $B? \to B$ and $B \to \varepsilon$, where
$\varepsilon$ is the empty word. Similarly, $B^*$ denotes a nonterminal with the production rules
$B^* \to B, B^*$ and $B^* \to \varepsilon$. To maintain determinism, one must ensure
two conditions: First, if a rule generates two adjacent nonterminals that are starred or
optional, then these nonterminals must be different, so $A \to x(B*, C*)$ is permitted but
$A \to x(B*, B?)$ is not, because we would not know whether to assign an element to $B*$ or
$B?$. Second, the languages generated by any two right-hand-sides for the same nonterminal must be non-intersecting.
For example, if the rule $A \to x(B*, C)$ exists, then the rule $A \to x(C?, D*)$ is not allowed because
the right-hand-side $x(C)$ could be generated by either of them (refer to Appendix~\ref{ap:deterministic} for more details).
In the remainder of this paper, we generally assume that we deal with deterministic regular
tree grammars that may contain starred and optional nonterminals.

\subsection{Tree Encoding}

We define a \emph{tree encoder} for a regular tree grammar $\mathcal{G}$ as a mapping
$\Enc : \mathcal{L}(\mathcal{G}) \to \R^n$ for some encoding dimensionality $n \in \N$.
While fixed tree encodings do exist, e.g.\ in the form of tree kernels \citep{Aiolli2015,Collins2002}, 
we focus here on \emph{learned} encodings via deep neural networks.
A simple tree encoding scheme is to list all nodes of a tree in depth-first-search order and encode this list via a recurrent
or convolutional neural network \citep{Paassen2020IJCNN}. However, one can also encode the tree
structure more directly via recursive neural networks 
\citep{Gallicchio2013,Tai2015,Pollack1990,Sperduti1997,Sperduti1994}.
Generally speaking, a recursive neural network consists of a set of mappings
$\enc^x : \mathcal{P}(\R^n) \to \R^n$, one for each symbol $x \in \Sigma$,
which receive a (perhaps ordered) set of child encodings as input and map it to a parent
encoding.
Based on such mappings, we define the overall tree encoder recursively as
\begin{equation}
\Enc\big( x(\tre y_1, \ldots, \tre y_k) \big) := \enc^x\big(\{ \Enc(y_1), \ldots, \Enc(y_k) \}\big).
\end{equation}
Traditional recursive neural networks implement $\enc^x$ with single- or multi-layer perceptrons.
More recently, recurrent neural networks have been applied, such as echo state nets
\citep{Gallicchio2013} or LSTMs \citep{Tai2015}. In this work, we extend the encoding scheme
by defining the mappings $\enc^x$ not over terminal symbols $x$ but over grammar rules $r$, thereby
tying encoding closely to parsing. This circumvents a typical problem in recursive neural nets,
namely to handle the order and number of children \citep{Sperduti1997}.

Recursive neural networks can also be related to more general graph neural networks
\citep{Kipf2017,Micheli2009,Scarselli2009}. In particular, we can interpret a recursive neural network
as a graph neural network which transmits messages from child nodes to parent nodes until the
root is reached. Thanks to the acyclic nature of trees, a single pass from
leaves to root is sufficient, whereas most graph neural net
architectures would require as many passes as the tree is deep \citep{Kipf2017,Micheli2009,Scarselli2009}.
In other words, graph neural nets only consider neighboring nodes in a pass, whereas recursive
nets incorporate information from all descendant nodes. Another reason why
we choose to consider trees instead of general graphs is that graph grammar
parsing is NP-hard \citep{Turan1983}, whereas regular tree grammar parsing is linear \citep{Comon2008}.

For the specific application of encoding syntax trees of computer programs, three further
strategies have been proposed recently, namely:
Code2vec considers paths from the root to single nodes and aggregates
information across these paths using attention \citep{Alon2019}; AST-NN treats a syntax tree as a
sequence of subtrees and encodes these subtrees first, followed by a GRU which encodes the sequence
of subtree encodings \citep{Zhang2019ASTNN};
and CuBERT treats source code as a sequence of tokens which are then plugged into a big transformer
model from natural language processing \citep{Kanade2020}.
Note that these models focus on encoding trees, whereas our focus lies on
decoding.

\subsection{Tree Decoding}

We define a \emph{tree decoder} for a regular tree grammar $\mathcal{G}$ as a mapping
$\Dec : \R^n \to \mathcal{L}(\mathcal{G})$ for some encoding dimensionality $n \in \N$.
In early work, \citet{Pollack1990} and \citet{Sperduti1994} already proposed
decoding mechanisms using 'inverted' recursive neural networks, i.e.\ mapping from a parent
representation to a fixed number of children, including a special 'none' token for missing children.
Theoretical limits of this approach have been investigated by \citet{Hammer2002}, who showed
that one requires exponentially many neurons to decode all possible trees of a certain depth.
More recently, multiple works have considered the more general problem of decoding graphs from vectors,
where a graph is generated by a sequence of node and edge insertions, which in turn is generated
via a deep recurrent neural net \citep{Zhang2019,Bacciu2019,Liu2018,Paassen2021ICLR,You2018}.
From this family, the variational autoencoder for directed acyclic graphs (D-VAE) \citep{Zhang2019}
is most suited to trees because it explicitly prevents cycles. In particular, the network
generates nodes one by one and then decides which of the earlier nodes to connect to the new
node, thereby preventing cycles. We note that there is an entire branch of graph generation devoted
specifically to molecule design which is beyond our capability to cover
here \citep{SanchezLengeling2018}. However, tree decoding may serve as a subroutine,
e.g.\ to construct a junction tree in \citep{Jin2018}.

Another thread of research concerns the generation of strings from a context-free grammar, guided
by a recurrent neural network \citep{Kusner2017,Dai2018}. Roughly speaking, these approaches
first parse the input string, yielding a generating rule sequence,
then convert this rule sequence into a vector via a convolutional neural net, and finally decode
the vector back into a rule sequence via a recurrent neural net. This rule sequence, then, yields
the output string.
Further, one can incorporate additional syntactic or semantic constraints via attribute grammars
in the rule selection step \citep{Dai2018}. We follow this line of research but use tree
instead of string grammars and employ recursive
instead of sequential processing. This latter change is key because it ensures that
the distance between the encoding and decoding of a node is bounded by the tree depth
instead of the tree size, thus decreasing the required memory capacity from linear to
logarithmic in the tree size.

A third thread of research attempts to go beyond known grammars and instead tries to
infer a grammar from data, typically using stochastic parsers and
grammars that are controlled by neural networks 
\citep{Allamanis2017,Dyer2016,Li2019,Kim2019,Yogatama2017,Zaremba2014}. Our
work is similar in that we also control a parser and a grammar with a neural network. However,
our task is conceptually different: We assume a grammar is given and are solely concerned with
autoencoding trees within the grammar's language, whereas these works attempt to find tree-like
structure in strings. While this decision constrains us to known grammars, it also enables us
to consider non-binary trees and variable-length rules which are currently beyond grammar induction
methods. Further, pre-specified grammars are typically designed to support interpretation and
semantic evaluation (e.g.\ via an objective function for optimization). Such an interpretation
is much more difficult for learned grammars.

Finally, we note that our own prior work \citep{Paassen2020IJCNN} already combines tree grammars
with recursive neural nets (in particular tree echo state networks \citep{Gallicchio2013}).
However, in this paper we combine such an architecture with an end-to-end-learned variational
autoencoder, thus guaranteeing a smooth latent space, a standard normal distribution in the latent
space, and smaller latent spaces. It also yields empirically superior results, as we see later
in the experiments.

\subsection{Variational Autoencoders}

An autoencoder is a combination of an encoder $\Enc$ and a decoder $\Dec$ that is trained 
to minimize some form of autoencoding error, i.e.\ some notion of
dissimilarity between an input $\tre x$ and its autoencoded version $\Dec(\Enc(\tre x))$.
In this paper, we consider the variational autoencoder (VAE) approach of
\citet{Kingma2019}, which augments the deterministic encoder and decoder
to probability distributions from which we can sample.
More precisely, we introduce a probability density $q_\Enc(\vec z \vert \tre x)$
for encoding $\tre x$ into a vector $\vec z$, and a probability
distribution $p_\Dec(\tre x\vert \vec z)$ for decoding $\vec z$ into
$\tre x$.

Now, let $\tre x_1, \ldots, \tre x_m$ be a training data set.
We train the autoencoder to minimize the loss:
\begin{equation}
\ell(\Enc, \Dec) =
\sum_{i=1}^m \mathbb{E}_{q_\Enc(z_i \vert \tre x_i)}\Big[-\log\big[p_\Dec\big( \tre x_i \big\vert \vec z_i \big)\big]\Big] + \beta \cdot \mathcal{D}_\mathrm{KL}(q_\Enc\|\mathcal{N}), \label{eq:vae}
\end{equation}
where $\mathcal{D}_\mathrm{KL}$ denotes the Kullback-Leibler divergence
between two probability densities and where $\mathcal{N}$ denotes the density
of the standard normal distribution. $\beta$ is a hyper-parameter to weigh
the influence of the second term, as suggested by \citet{Burda2016}.

Typically, the loss in~\eqref{eq:vae} is minimized over tens of thousands
of stochastic gradient descent iterations, such that the expected value
over $q_\Enc(z_i \vert \tre x_i)$ can be replaced with a single sample
\citep{Kingma2019}. Further, $q_\Enc$ is typically modeled as a Gaussian
with diagonal covariance matrix, such that the sample can be re-written
as $\vec z_i = \rho(\vec \mu_i + \vec \epsilon_i \odot \vec \sigma_i)$,
where $\vec \mu_i$ and $\vec \sigma_i$ are deterministically generated
by the encoder $\Enc$, where $\odot$ denotes element-wise multiplication,
and where $\vec \epsilon_i$ is Gaussian noise, sampled with mean zero and
standard deviation $s$. $s$ is a hyper-parameter which regulates the noise
strength we impose during training.

We note that many extensions to variational autoencoders have been proposed
over the years \citep{Kingma2019}, such as Ladder-VAE \citep{Sonderby2016} or InfoVAE \citep{Zhao2019}.
Our approach is generally compatible with such extensions, but our focus here
lies on the combination of autoencoding, grammatical knowledge, and
recursive processing, such that we leave extensions of the autoencoding
scheme for future work.

\section{Method}

\begin{figure*}
\centering
\includegraphics[width=\textwidth]{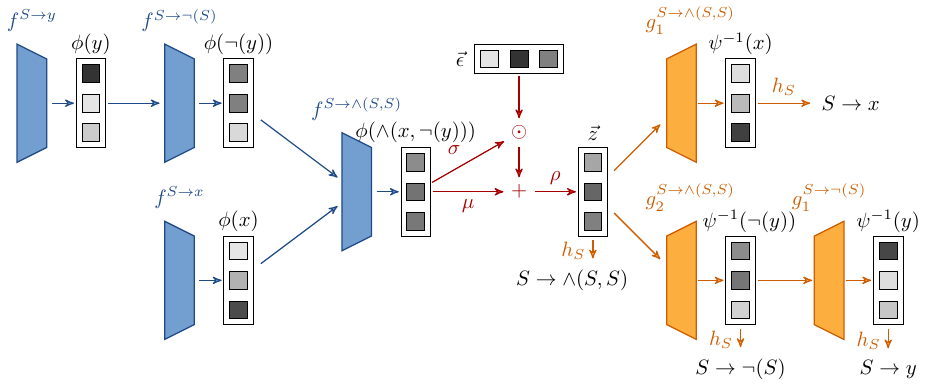}
\caption{An illustration of recursive tree grammar autoencoders (RTG-AEs) for the Boolean formula
$\tre x = \wedge(x, \neg(y))$, which is first encoded (left, blue) as a vector $\Enc(\tre x)$,
then equipped with Gaussian random noise $\vec \epsilon$ via the formula
$\vec z = \rho\big(\mu[\Enc(\tre x)] + \vec \epsilon \odot \sigma[\Enc(\tre x)]\big)$
(center, red), where$\odot$ is the element-wise product, and finally is decoded back
(right, orange) to a tree $\Dec(\vec z)$.
Encoding (Algorithm~\ref{alg:encoding}) and decoding (Algorithm~\ref{alg:decoding}) are
recursive parsing/production processes using the regular tree grammar
$\mathcal{G} = \big( \{ S\}, \{ \wedge, \vee, \neg, x, y\},
\{ S \to \wedge(S, S), S \to \vee(S, S), S \to \neg(S), S \to x, S \to y\}, S \big)$.}
\label{fig:approach}
\end{figure*}

Our proposed architecture is a variational autoencoder for trees, where we construct the encoder
as a bottom-up parser, the decoder as a regular tree grammar, and the reconstruction loss as
the crossentropy between the true rules generating the input tree and the rules
chosen by the decoder. An example autoencoding computation is shown in Figure~\ref{fig:approach}.
Because our encoding and decoding schemes are closely related to
recursive neural networks \citep{Pollack1990,Sperduti1997,Sperduti1994},
we call our approach \emph{recursive tree grammar autoencoders} (RTG-AEs). We now introduce each of the components in turn.

\subsection{Encoder}

Our encoder is a bottom-up parser for a given regular tree grammar
$\mathcal{G} = (\Phi, \Sigma, R, S)$, computing a vectorial representation in
parallel to parsing.
In more detail, we introduce an encoding function
$\enc^r : \R^{k \times n} \to \R^n$ for each grammar rule $r = (A \to x(B_1, \ldots, B_k)) \in R$,
which maps the encodings of all children to an encoding of the parent node. Here, $n$ is the
encoding dimensionality. As such, the grammar guides our encoding and fixes the
number and order of inputs for our encoding functions $\enc^r$. Note that, if $k = 0$,
$\enc^r$ is a constant. 

Next, we apply the functions $\enc^r$ recursively during parsing, yielding a vectorial encoding
of the overall tree. More precisely, our encoder $\Enc$ is defined by the
recursive equation
\begin{equation}
\Enc\Big(x(\tre y_1, \ldots, \tre y_K), A\Big) = 
\enc^r(\Enc(\tre y_1, B_1), \ldots, \Enc(\tre y_1, B_k)),
\end{equation}
where $r = A \to x(B_1, \ldots, B_k)$ is the first rule in the sequence
that generates $x(\tre y_1, \ldots, \tre y_K)$ from $A$. As initial nonterminal
$A$, we use the grammar's starting symbol $S$.
Refer to Algorithm~\ref{alg:encoding} for details.

We implement $\enc^r$ as a single-layer feedforward
neural network of the form $\enc^r(\vec y_1, \ldots, \vec y_k) = \tanh(\sum_{j=1}^k \bm{U}^{r, j} \cdot \vec y_j + \vec a^r )$,
where the weight matrices $\bm{U}^{r, j} \in \R^{n \times n}$ and the bias vectors
$\vec a^r \in \R^n$ are parameters to be learned.
For optional and starred nonterminals, we further define $\enc^{B? \to \varepsilon} =
\enc^{B^* \to \varepsilon} = \vec 0$, $\enc^{B? \to B}(\vec y) = \vec y$, and
$\enc^{B^* \to B, B^*}(\vec y_1, \vec y_2) = \vec y_1 + \vec y_2$. In other words,
the empty string $\varepsilon$ is encoded as the zero vector, an optional nonterminal is encoded via the identity,
and starred nonterminals are encoded via a sum, following the recommendation of
\citet{Xu2018} for graph neural nets.

\begin{algorithm}
\caption{Our encoding scheme, as first proposed in \citep{Paassen2020IJCNN}, which extends bottom-up parsing
for a deterministic regular tree grammar $\mathcal{G} = (\Phi, \Sigma, R, S)$.
For each rule $r \in R$ we assume an encoding function $\enc^r$.
The algorithm receives a tree as input and returns a nonterminal symbol, a rule sequence
that generates the tree from that nonterminal symbol, and a vectorial encoding.}
\label{alg:encoding}
\begin{algorithmic}[1]
\Function{encode}{a tree $\tre x = x(\tre y_1, \ldots, \tre y_k)$}
\For{$j \in \{1, \ldots, k\}$}
\State $B_j, \bar r_j, \vec y_j \gets $ \Call{encode}{$\tre y_j$}.
\EndFor
\If{$\exists A \in \Phi : A \to x(B_1, \ldots, B_k) \in R$}
\State $r \gets \big(A \to x(B_1, \ldots, B_k)\big)$.
\State \Return $A, ( r, \bar r_1, \ldots, \bar r_k), \enc^r(\vec y_1, \ldots, \vec y_k)$.
\Else
\State Error; $\tre x$ is not in $\mathcal{L}(\mathcal{G})$.
\EndIf
\EndFunction
\end{algorithmic}
\end{algorithm}

We can show that Algorithm~\ref{alg:encoding} returns without error if and only if the input
tree is part of the grammar's tree language.

\begin{thm} \label{thm:parsing}
Let $\mathcal{G} = (\Phi, \Sigma, R, S)$ be a deterministic regular tree grammar.
Then, it holds: $\tre x$ is a tree in $\mathcal{L}(\mathcal{G})$ if and only if
Algorithm~\ref{alg:encoding} returns the nonterminal $S$ as first output.
Further, if Algorithm~\ref{alg:encoding} returns with $S$ as first output and some rule
sequence $\bar r$ as second output, then $\bar r$ uniquely generates $\tre x$ from $S$.
Finally, Algorithm~\ref{alg:encoding} has $\mathcal{O}(\lvert \tre x\rvert )$ time and space complexity.

\begin{proof}
Refer to Appendix~\ref{ap:parsing}.
\end{proof}
\end{thm}

\subsection{Decoder}

Our decoder is a stochastic version of a given regular tree grammar
$\mathcal{G} = (\Phi, \Sigma, R, S)$, controlled by two
kinds of neural network. First, for any nonterminal $A \in \Phi$, let $L_A$ be the number of rules in $R$ with $A$ on the left hand side.
For each $A \in \Phi$, we introduce a linear layer $\out_A : \R^n \to \R^{L_A}$
with $\out_A(\vec x) = \bm{V}^A \cdot \vec x + \vec b^A$.
To decode a tree from a vector $\vec x \in \R^n$ and a nonterminal $A \in \Phi$, we first
compute rule scores $\vec \lambda = \out_A(\vec x)$ and then sample a rule
$r_l = (A \to x(B_1, \ldots, B_k)) \in R$ from the softmax distribution
$p_A(r_l\vert\vec x) = \exp(\lambda_l) / \sum_{l' = 1}^{L_A} \exp(\lambda_{l'})$.
Then, we apply the sampled rule and use a second kind of neural network to decide the vectorial
encodings for each generated child nonterminal $B_1, \ldots, B_k$. In particular,
for each grammar rule $r = (A \to x(B_1, \ldots, B_k)) \in R$, we introduce $k$ 
decoding functions $\dec^r_1, \ldots, \dec^r_k : \R^n \to \R^n$ and compute the vector
encoding for the $j$th child as $\vec y_j = \dec^r_j(\vec x)$. Finally, we decode the children
recursively until no nonterminal is left.
More precisely, the tree decoding is guided by the recursive equation
\begin{equation}
\Dec(\vec x, A) = x\Big(\Dec(\vec y_1, B_1), \ldots, \Dec(\vec y_k, B_k) \Big),
\end{equation}
where the rule $r = x \to A(B_1, \ldots, B_k)$ is sampled from $p_A$ as specified
above and $\vec y_j = \dec^r_j(\vec x)$. As initial nonterminal argument we
use the grammar's starting symbol $S$.
For details, refer to 
Algorithm~\ref{alg:decoding}. Note that the time and space complexity is $\mathcal{O}(\lvert \tre x\rvert )$
for output tree $\tre x$ because each recursion step adds exactly one terminal symbol.
Since the entire tree needs to be stored, the space complexity is
also $\mathcal{O}(\lvert \tre x\rvert )$. Also note that Algorithm~\ref{alg:decoding} is not generally
guaranteed to halt \citep{Chi1999}. In practice, we solve this problem by imposing a 
maximum number of generated rules.

\begin{algorithm}
\caption{Our decoding scheme, similar to the scheme in \citep{Paassen2020IJCNN}, which samples a rule
sequence from a regular tree grammar $\mathcal{G} = (\Phi, \Sigma, R, S)$. The algorithm receives a
vector $\vec x$ and the starting nonterminal $S$ as input and then samples rules with probabilities
determined by functions $\out_A$ for each nonterminal $A \in \Phi$ and decoding functions $\dec^r_j$
for each rule $r = A \to x(B_1, \ldots B_k) \in R$.}
\label{alg:decoding}
\begin{algorithmic}[1]
\Function{decode}{vector $\vec x \in \R^n$, nonterminal $A \in \Phi$}
\State Let $r_1, \ldots, r_{L_A} \in R$ be all rules with $A$ on the left hand side.
\State Compute softmax weights $\vec \lambda \gets \out_A(\vec x)$.
\State Sample $r_l$ with probability
\State $\quad p_A(r_l\vert\vec x) = \exp(\lambda_l) / \sum_{l'=1}^{L_A} \exp(\lambda_{l'})$.
\State Let $r_l = A \to x(B_1, \ldots, B_k)$.
\For{$j \in \{1, \ldots, k\}$}
\State $\vec y_j \gets \dec^{r_l}_j(\vec x)$.
\State $\tre y_j \gets$ \Call{decode}{$\vec y_j$, $B_j$}.
\State $\vec x \gets \vec x - \vec y_j$.
\EndFor
\State \Return $x(\tre y_1, \ldots, \tre y_k)$.
\EndFunction
\end{algorithmic}
\end{algorithm}

An interesting special case are trees that implement lists. For example, consider
a carbon chain CCC from chemistry. In the SMILES grammar \citep{Weininger1988}, this is represented as a
binary tree of the form single\_chain(single\_chain(single\_chain(chain\_end, C), C), C),
i.e.\ the symbol `single\_chain' acts as a list operator. In such a case, we recommend to
use a recurrent neural network to implement the decoding function $\dec^{\text{Chain} \to \text{single\_chain}(\text{Chain}, \text{Atom})}_1$,
such as a gated recurrent unit (GRU) \citep{Cho2014}. In all other cases, we stick with a simple
feedforward layer.

\subsection{Training}

\begin{algorithm}
\caption{A scheme to compute the variational autoencoder (VAE) loss from Equation~\ref{eq:vae}
for our approach. The algorithm receives a tree $\tre x$ as input and returns the VAE loss,
according to a mean function $\mu$, a standard deviation function $\sigma$, a VAE decoding layer
$\rho$, decoding functions $\dec^r_j : \R^n \to \R^n$, and VAE hyperparameters $s, \beta \in \R^+$.}
\label{alg:loss}
\begin{algorithmic}[1]
\Function{loss}{a tree $\tre x$}
\State $A, ( r_1, \ldots, r_T), \vec x \gets$ \Call{encode}{$\tre x$} via Algorithm~\ref{alg:encoding}.
\State $\vec \mu \gets \mu(\vec x)$. $\quad \vec \sigma \gets \sigma(\vec x)$.
$\quad \vec \epsilon \sim \mathcal{N}(\vec 0 \vert s \cdot \bm{I}^{n_\mathrm{VAE}})$.
\State $\vec z \gets \rho(\vec \mu + \vec \epsilon \odot \vec \sigma)$.
\State Initialize a stack $\mathcal{S}$ with $\vec z$ on top.
\State Initialize $\ell \gets \beta \cdot \big(\sum_{j=1}^{n_\mathrm{VAE}}
\mu_j^2 + \sigma_j^2 - \log[\sigma_j^2] - 1\big)$.
\For{$t \gets 1, \ldots, T$}
	\State Let $r_t = A \to x(B_1, \ldots, B_k)$.
	\State Pop $\vec x_t$ from the top of $\mathcal{S}$.
	\State Compute $p_A(r_t\vert\vec x_t)$ as in line 5 of Algorithm~\ref{alg:decoding}.
	\State $\ell \gets \ell - \log\big[ p_A(r_t \vert \vec x_t) \big]$.
	\For{$j \gets 1, \ldots, k$}
		\State $\vec y_j \gets \dec^{r_t}_j(\vec x_t)$.
		\State $\vec x_t \gets \vec x_t - \vec y_j$.
	\EndFor
	\State Push $\vec y_k, \ldots, \vec y_1$ onto $\mathcal{S}$.
\EndFor
\State \Return $\ell$.
\EndFunction
\end{algorithmic}
\end{algorithm}

We train our recursive tree grammar autoencoder (RTG-AE) in the variational autoencoder (VAE)
framework, i.e.\ we try to minimize the loss in Equation~\ref{eq:vae}.
More precisely, we define the encoding probability density $q_\Enc(\vec z\vert\tre x)$
as the Gaussian with mean $\mu(\Enc(\tre x))$ and covariance matrix
$\mathrm{diag}\big[\sigma(\Enc(\tre x))\big]$, where the
functions $\mu : \R^n \to \R^{n_\mathrm{VAE}}$ and
$\sigma : \R^n \to \R^{n_\mathrm{VAE}}$ are defined as
\begin{align}
\mu(\vec x) &= \bm{U}^\mu \cdot \vec x + \vec a^\mu, \notag \\
\sigma(\vec x) &= \exp\big(\frac{1}{2} \cdot [\bm{U}^\sigma \cdot \vec x + \vec a^\sigma]\big), \label{eq:vae_funs}
\end{align}
where $\bm{U}^\mu, \bm{U}^\sigma \in \R^{n_\mathrm{VAE} \times n}$ and $\vec a^\mu, \vec a^\sigma \in \R^{n_\mathrm{VAE}}$ are additional parameters.

To decode, we first transform the encoding vector $\vec z$ with a single
layer $\rho(\vec z) = \tanh\big(\bm{W}^\rho \cdot \vec z + \vec c^\rho\big)$
and then apply the decoding scheme from Algorithm~\ref{alg:decoding}.
As the decoding probability $p_\Dec(\tre x\vert\vec z)$,
we use the product over all probabilities $p_A(r_t\vert\vec x_t)$ from line 5 of Algorithm~\ref{alg:decoding},
i.e.\ the probability of always choosing the correct grammar rule during decoding, provided that
all previous choices have already been correct. The negative logarithm of this product can also
be interpreted as the crossentropy loss between the correct rule sequence and the softmax probabilities
from line 5 of Algorithm~\ref{alg:decoding}.
The details of our loss computation are given in Algorithm~\ref{alg:loss}. Note that the time
and space complexity is $\mathcal{O}(\lvert \tre x\rvert )$ because the outer loop from line 7-17 runs
$T = \lvert \tre x\rvert $ times, and the inner loop in lines 12-15 runs $\lvert \tre x\rvert -1$ times in total
because every node takes the role of child exactly once (except for the root).
Because the loss is differentiable, we can optimize it using gradient
descent schemes such as Adam \citep{Adam2013}. The gradient computation is performed by
the pyTorch autograd system \citep{pytorch2019}.

\section{Experiments and Discussion}

\begin{table}
\centering
\begin{tabular}{lcccc}
& Boolean & Expressions & SMILES & Pysort \\
\cmidrule(lr){1-1} \cmidrule(lr){2-5}
dataset size & $34,884$ & $104,832$ & $249,456$ & $294$ \\
avg.\ tree size & $5.21$ & $8.95$ & $83.66$ & $57.97$ \\
avg.\ depth & $3.18$ & $4.32$ & $21.29$ & $9.25$ \\
max.\ rule seq.\ length & $14$ & $11$ & $285$ & $168$ \\
no.\ symbols & $5$ & $9$ & $43$ & $54$ \\
no.\ grammar rules & $5$ & $9$ & $46$ & $54$
\end{tabular}
\caption{The statistics for all four dataset.}
\label{tab:data}
\end{table}

\begin{table}
\centering
\begin{tabular}{lcccc}
model & Boolean & Expressions & SMILES & Pysort\\
\cmidrule(lr){1-1} \cmidrule(lr){2-5}
D-VAE & $249,624$ & $252,828$ & $1,979,038$ & $1,859,417$ \\
GVAE & $198,316$ & $197,536$ & $1,928,431$ & $1,630,263$ \\
TES-AE & $500$ & $900$ & $11,776$ & $13,824$ \\
GRU-TG-AE & $67,221$ & $70,025$ & $437,080$ & $445,280$ \\
\cmidrule(lr){1-1} \cmidrule(lr){2-5}
RTG-AE & $104,021$ & $165,125$ & $5,161,816$ & $5,493,600$ \\
\end{tabular}
\caption{The number of parameters for all models on all datasets.}
\label{tab:parameters}
\end{table}

We evaluate the performance of RTG-AEs on four datasets (refer to Table~\ref{tab:data}
for statistics), namely:

\textbf{Boolean:} Randomly sampled Boolean formulae over the variables $x$ and $y$
with at most three binary operators, e.g.\ $x \wedge \neg y$ or $x \wedge y \wedge (x \vee \neg y)$.

\textbf{Expressions:} Randomly sampled algebraic expressions over the variable $x$
of the form $3 * x + \sin(x) + \exp(2 / x)$, i.e.\ consisting of a binary
operator plus a unary operator plus a unary of a binary. This dataset is taken from
\citep{Kusner2017}.

\textbf{SMILES:} Roughly 250k chemical molecules as SMILES strings
\citep{Weininger1988}, as selected by \citep{Kusner2017}.

\textbf{Pysort:} 29 Python sorting programs and manually generated preliminary
development stages of these programs, resulting in 294 programs overall.

\begin{table}
\centering
\begin{tabular}{lcccc}
model & Boolean & Expressions & SMILES & Pysort\\
\cmidrule(lr){1-1} \cmidrule(lr){2-5}
D-VAE & $4.32 \pm 0.41$ & $5.84 \pm 0.32$ & $132.70 \pm 57.02$ & $70.67 \pm 7.20$ \\
GVAE & $3.61 \pm 0.51$ & $5.84 \pm 2.00$ & $594.92 \pm 5.99$ & $61.44 \pm 5.18$ \\
TES-AE & $2.62 \pm 0.26$ & $2.09 \pm 0.18$ & $581.08 \pm 25.99$ & $\bm{20.40 \pm 4.27}$ \\
GRU-TG-AE & $1.98 \pm 0.53$ & $3.70 \pm 0.31$ & $482.88 \pm 129.52$ & $93.63 \pm 1.44$ \\
\cmidrule(lr){1-1} \cmidrule(lr){2-5}
RTG-AE & $\bm{0.83 \pm 0.23}$ & $\bm{0.77 \pm 0.23}$ & $\bm{111.14 \pm 159.97}$ & $38.14 \pm 3.63$ \\
\end{tabular}
\caption{The average autoencoding RMSE ($\pm$ standard deviation).}
\label{tab:rmses}
\end{table}

\begin{table}
\centering
\begin{tabular}{lcccc}
model & Boolean & Expressions & SMILES & Pysort\\
\cmidrule(lr){1-1} \cmidrule(lr){2-5}
D-VAE & $882.0 \pm 45.9$ & $1201.5 \pm 24.8$ & $64225.9 \pm 4,500.3$ & $66789.9 \pm 3555.2$ \\
GVAE & $1316.5 \pm 86.7$ & $1218.8 \pm 73.8$ & $7771.2 \pm 1,524.2$ & $1393.7 \pm 120.9$ \\
TES-AE & $\bm{0.9 \pm 0.1}$ & $\bm{1.5 \pm 0.2}$ & $\bm{583.6 \pm 103.6}$ & $\bm{10.1 \pm 0.6}$ \\
GRU-TG-AE & $440.8 \pm 30.4$ & $734.5 \pm 50.8$ & $3342.2 \pm 42.7$ & $648.6 \pm 50.1$ \\
\cmidrule(lr){1-1} \cmidrule(lr){2-5}
RTG-AE & $221.5 \pm 4.7$ & $357.4 \pm 13.2$ & $1903.4 \pm 34.1$ & $399.9 \pm 21.9$ \\
\end{tabular}
\caption{The average runtime in seconds ($\pm$ standard deviation) as measured by Python.}
\label{tab:runtimes}
\end{table}

We compare RTG-AEs to three baseline models from the literature, namely grammar variational
autoencoders (GVAE) \citep{Kusner2017}, directed acyclic graph variational autoencoders (D-VAE)
\citep{Zhang2019}, and tree echo state autoencoders (TES-AE) \citep{Paassen2020IJCNN}.
Additionally, we compare to a recurrent version of tree grammar AEs (GRU-TG-AEs), i.e.\
we represent a tree by its rule sequence and autoencode it with a gated recurrent unit (GRU) \citep{Cho2014}.
We also performed preliminary experiments with semantic constraints akin to \citep{Dai2018}
but did not obtain significantly different results, such that we omit such constraints here for
simplicity.
Note that our baselines are selected to implement an ablation study: GVAE and GRU-TG-AE use grammar knowledge and variational autoencoding (VAE), but not recursive processing;
TES-AE uses grammar knowledge and recursive processing, but not VAE; and D-VAE uses
graph-like processing and VAE but no grammar knowledge.

We used the reference implementation and architecture of all approaches, with a slight
difference for GVAE, where we ported the reference implementation from Keras to pyTorch in order
to be comparable to all other approaches. The number of parameters for all models on all
datasets is shown in Table~\ref{tab:parameters}.
We trained all neural networks using Adam \citep{Adam2013} with a learning rate of $10^{-3}$ and a 
\texttt{ReduceLROnPlateau} scheduler with minimum learning rate $10^{-4}$, following the
learning procedure of \citet{Kusner2017} as well as \citet{Dai2018}. We sampled 100k trees for the first two and 10k
trees for the latter two datasets.
To obtain statistics, we repeated the training ten times with different samples (by generating new data for boolean
and expressions and by doing a 10-fold crossvalidation for SMILES and pysort).
Following \citet{Zhang2019}, we used a batch size of $32$ for all approaches%
\footnote{On the Pysort dataset, the batch size of D-VAE had to be reduced to $16$ to avoid memory overload.}.
For each approach and each dataset, we optimized the hyper-parameters $\beta$ and $s$ of Equations~\ref{eq:vae}
and~\ref{eq:vae_funs} in a random search with 20 trials over the range $[10^{-5}, 1]$, using separate data. For the first two data sets, we set $n = 100$ and $n_\mathrm{VAE} = 8$,
whereas for the latter two we set $n = 256$ and $n_\mathrm{VAE} = 16$. For TES-AE, we followed
the protocol of \citet{Paassen2020IJCNN}, training on a random subset of $500$ training data points,
and we optimized the sparsity, spectral radius, regularization strength with the same hyper-parameter optimization scheme.

\begin{figure*}[t]
\centering
\includegraphics[width=\textwidth]{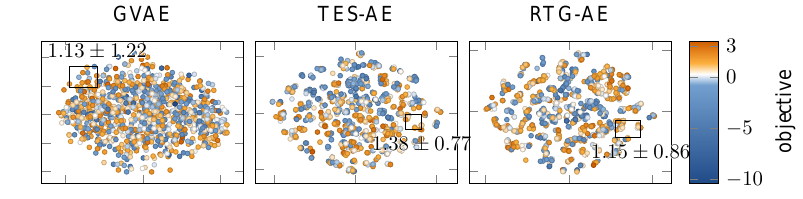}
\caption{A 2D t-SNE reduction of the codes of 1000 random molecules
from the SMILES data set by GVAE (left), TES-AE (center), and RTG-AE (right).
Color indicates the objective function value. Rectangles indicate the $30$-means cluster
with highest mean objective function value $\pm$ std.}
\label{fig:coding_space}
\end{figure*}

\begin{table}
\centering
\begin{tabular}{lcccc}
model & Boolean & Expressions & SMILES & Pysort \\
\cmidrule(lr){1-1} \cmidrule(lr){2-5}
D-VAE & $28.8\%$ & $23.8\%$ & $0.1\%$ & $1.4\%$\\
GVAE & $90.4\%$ & $99.8\%$ & $0\%$ & $99.9\%$\\
GRU-TG-AE & $99.9\%$ & $98.2\%$ & $5.9\%$ & $0\%$\\
\cmidrule(lr){1-1} \cmidrule(lr){2-5}
RTG-AE & $98.5\%$ & $99.6\%$ & $37.3\%$ & $99.7\%$
\end{tabular}
\caption{The rate of syntactically correct trees decoded from standard normal random vectors.}
\label{tab:syntactic_correctness}
\end{table}

\begin{table}
\centering
\begin{tabular}{lcc}
model & median tree & median score \\
\cmidrule(lr){1-1} \cmidrule(lr){2-2} \cmidrule(lr){3-3}
& Expressions & \\
\cmidrule(lr){1-1} \cmidrule(lr){2-2} \cmidrule(lr){3-3}
D-VAE & x & $0.49 \pm 0.00$ \\
GVAE & x + 1 + sin(3 + 3) & $0.46 \pm 0.00$ \\
TES-AE & x & $0.49 \pm 0.00$ \\
GRU-TG-AE & x & $0.49 \pm 0.00$ \\
RTG-AE & x + 1 + sin(x * x) & $\bm{0.33 \pm 0.05}$ \\
\cmidrule(lr){1-1} \cmidrule(lr){2-2} \cmidrule(lr){3-3}
& SMILES & \\
\cmidrule(lr){1-1} \cmidrule(lr){2-2} \cmidrule(lr){3-3}
GVAE & n.a. & n.a. \\
TES-AE & CCO & $-0.31 \pm 0.50$ \\
GRU-TG-AE & C=C & $-2.23 \pm 1.56$ \\
RTG-AE & CCCCCCC & $\bm{2.57 \pm 0.31}$ \\
\end{tabular}
\caption{The optimization scores for the Expressions (lower is better) and SMILES
(higher is better) with median tree and median score ($\pm \frac{1}{2}$ IQR).}
\label{tab:optimization}
\end{table}

The SMILES experiment was performed on a computation server with a 24core CPU and 48GB RAM,
whereas all other experiments were performed on a consumer grade laptop with Intel i7 4core CPU
and 16GB RAM.
All experimental code, including all grammars and implementations, is available at
\url{https://gitlab.com/bpaassen/rtgae}.

We measure autoencoding error on test data in terms of the root mean square
tree edit distance \citep{Zhang1989}. We use the root mean square error (RMSE) rather than log
likelihood because D-VAE measures log likelihood different to GVAE, GRU-TG-AE, and RTG-AE, and TES-AE
does not measure log likelihood at all. By contrast, the RMSE is agnostic to the underlying
distribution model. Further, we use the tree edit distance as a tree metric because it is defined
on all possible labeled trees without regard for the underlying distribution or grammar and hence
does not favor any of the models \citep{Bille2005}.

The RMSE results are shown in Table~\ref{tab:rmses}. We observe that RTG-AEs achieve notably
lower errors compared to all baselines on all datasets ($p < 0.05$ in a Wilcoxon signed rank test,
except for D-VAE on SMILES, where $p > 0.05$, and TES-AE on pysort, where RTG-AEs were worse).
We also note that, among all deep learning approaches, RTG-AEs have the lowest training times,
speeding up at least one third over the fastest other baseline, GRU-TG-AEs (refer to
Table~\ref{tab:runtimes}). Only TES-AEs are notably faster because they do not require
backpropagation. We explain these results by the fact that all three defining features of RTG-AEs
contribute to a low error: Compared to D-VAEs, RTG-AEs exploit the inductive bias provided by a
regular tree grammar and can, hence, achieve a lower error with less training. Compared to GVAEs
and GRU-TG-AEs, RTG-AEs use recursive processing, which means that information only needs to be
remembered across the depth of a tree, not across the number of nodes. This means that a smaller model 
suffices to achieve the desired memory capacity. Finally, compared to TES-AEs, RTG-AEs train the
autoencoder end-to-end in a variational autoencoder framework which gives the model much more ability
to adjust to the data set at hand. However, we notice that these additional degrees of freedom only
help if the data set is large enough: For pysort, by far the smallest data set, TES-AEs do better.

To evaluate the ability of all models to generate syntactically valid trees, we sampled 1000
standard normal random vectors and decoded them with all models\footnote{We excluded TES-AEs in
this analysis because they do not guarantee a Gaussian distribution in the latent space and, hence,
are not compatible with this sampling approach.}. Then, we checked their syntax
with a bottom-up regular tree parser for the language. The percentage of correct trees is shown
in Table~\ref{tab:syntactic_correctness}. Unsurprisingly, D-VAE has the worst results across the
board because it does not use grammatical knowledge for decoding. In principle, the other models
should always have 100\% because their architecture guarantees syntactic correctness. Instead, we
observe that the rates drop far below 100\% for GRU-TG-AE on Pysort and for all models on SMILES.
This is because the decoding process can fail if it gets stuck in a loop. Even on the SMILES
dataset, though, the RTG-AE achieves the best rate.

We also repeated the optimization experiments of \citet{Kusner2017} in the latent space of 
the Expressions and SMILES datasets. We used the same objective functions as \citet{Kusner2017};
i.e.\ the log MSE to the ground truth expression $\frac{1}{3} + x + \sin(x * x)$ for Expressions
and a mixture of logP, synthetic availability, and cycle length for SMILES. For SMILES,
we further retrained all models (except D-VAE due to memory constraints) with $n_\mathrm{VAE} = 56$
and 50k training samples to have the same model capacity and training set size as \citet{Kusner2017}.
In contrast to \citet{Kusner2017}, we used CMA-ES instead of Bayesian
optimization because we could not get the original Bayesian Optimization implementation
to run within reasonable effort, whereas the Python cma package worked out-of-the-box.
CMA-ES is a usual method for high-dimensional, gradient-free optimization that has shown
competitive results to Bayesian optimization for neural network optimization \citep{Loshchilov2016}.
It also is particularly well suited to the latent space of VAEs because CMA-ES samples its
population from a Gaussian distribution.
We used $15$ iterations and a computational budget of $750$
trees, as \citet{Kusner2017}.
To obtain statistics, we performed the optimization 10 times for each approach.

The median results ($\pm$ inter-quartile ranges) are shown in Table~\ref{tab:optimization}.
We observe that RTG-AEs significantly outperform all baselines on both data sets ($p < 0.01$ in a Wilcoxon 
rank-sum test) by a difference of several inter-quartile ranges.
For SMILES, CMA-ES couldn't find any semantically valid molecule for GVAE and
achieved a negative median score for all methods but RTG-AE. We note that, on both
data sets, Bayesian optimization performed better than CMA-ES \citep{Kusner2017}.
Still, our results show that even the weaker CMA-ES optimizer can consistently
achieve good scores in the RTG-AE latent space. We believe there are two reasons for this:
First, the higher rate of syntactically correct trees for RTG-AE (refer to Table~\ref{tab:syntactic_correctness});
second, because recursive processing tends to cluster similar trees together \citep{Paassen2020IJCNN,Tino2003} in a fractal fashion,
such that an optimizer only needs to find a viable cluster and optimize within it.
Figure~\ref{fig:coding_space} shows a t-SNE visualization of the latent spaces, revealing that the
recursive models provide more distinct clusters, compared to GVAE, and that trees inside the cluster with
best objective function value have lower variance compared to GVAE.

\section{Conclusion}

In this contribution, we introduced the recursive tree grammar autoencoder (RTG-AE), a novel neural network architecture that combines
variational autoencoders with recursive neural networks and regular tree grammars.
In particular, our approach encodes a tree with a bottom-up parser,
and decodes it with a tree grammar, both learned via neural networks and variational autoencoding.
In an ablation study, we showed that the unique combination of recursive processing, grammatical knowledge, and variational autoencoding 
improves autoencoding error, training time, and optimization performance beyond
existing models that use only two of these concepts, but not all three.
This finding can be explained by three conceptual observations: First,
recursive processing follows the tree structure whereas sequential processing
introduces long-range dependencies between children and parents in the tree;
second, grammatical knowledge avoids obvious decoding mistakes by limiting the
terminal symbols we can choose; third, variational autoencoding encourages a
smooth encoding space, whereas a random encoding may exhibit unfavourable
structure for sampling and optimization. Theoretically, we proved that RTG-AEs
parse and generate trees in linear time and are expressive enough for all
regular tree languages.

In future work, one could replace more encoders and decoders with recurrent networks,
such as Tree-LSTMs, and one could impose more domain-specific semantic constraints as well as
post-processing mechanisms to tailor RTG-AEs to specific application domains.

\section*{Acknowledgment}

Funding by the German Research Foundation (DFG) under grant number PA 3460/1-1
is gratefully acknowledged.

\bibliographystyle{plainnat}
\bibliography{literature}

\begin{appendix}
	
\section{Proofs}

For the purpose of our tree language proofs, we first introduce a few auxiliary concepts.
First, we extend the definition of a regular tree grammar slightly to
permit multiple starting symbols. In particular, we re-define a regular tree grammar as a
$4$-tuple $\mathcal{G} = (\Phi, \Sigma, R, \mathcal{S})$, where $\Phi$ and $\Sigma$ are
disjoint finite sets and $R$ is a rule set as before, but $\mathcal{S}$ is now a subset of
$\Phi$. Next, we define the partial tree language $\mathcal{L}(\mathcal{G}\vert A)$ for nonterminal
$A \in \Phi$ as the set of all trees which can be generated from nonterminal $A$ using
rule sequences from $R$. Further, we define the $n$-restricted partial language
$\mathcal{L}_n(\mathcal{G}\vert A)$ for nonterminal $A \in \Phi$ as the set
$\mathcal{L}_n(\mathcal{G}\vert A) := \big\{ \tre{x} \in \mathcal{L}(\mathcal{G}\vert A) \big\vert
\vert \tre{x}\vert  \leq n \big\}$, i.e.\ it includes only the trees up to size $n$.
We define the tree language $\mathcal{L}(\mathcal{G})$ of $\mathcal{G}$ as the
union $\mathcal{L}(\mathcal{G}) := \bigcup_{S \in \mathcal{S}} \mathcal{L}(\mathcal{G}\vert S)$.

As a final preparation, we introduce a straightforward lemma regarding restricted partial
tree languages.

\begin{lma}\label{lma:grammar}
Let $\mathcal{G} = (\Phi, \Sigma, R, \mathcal{S})$ be a regular tree grammar.

Then, for all $A \in \Phi$, all $n \in \N$ and all trees $\tre{x}$ it holds:
$\tre{x} = x(\tre{y}_1, \ldots, \tre{y}_k)$ is in $\mathcal{L}_n(\mathcal{G}\vert A)$ if and
only if there exist nonterminals $B_1, \ldots, B_k \in \Phi$ such that
$A \to x(B_1, \ldots, B_k) \in R$ and $\tre{y}_j \in \mathcal{L}_{n_j}(\mathcal{G}\vert B_j)$
for all $j \in \{1, \ldots, k\}$ with $n_1 + \ldots + n_k = n-1$.

\begin{proof}
If $\tre{x} \in \mathcal{L}_n(\mathcal{G}\vert A)$, then there exists a sequence of rules
$r_1, \ldots, r_T \in R$ which generates $\tre{x}$ from $A$. The first rule in that sequence
must be of the form $A \to x(B_1, \ldots, B_k)$, otherwise $\tre{x}$ would not have the
shape $\tre{x} = x(\tre{y}_1, \ldots, \tre{y}_k)$. Further, $r_2, \ldots, r_T$ must
consist of subsequences $\bar r_1, \ldots, \bar r_k = r_2, \ldots, r_T$ such that
$\bar r_j$ generates $\tre{y}_j$ from $B_j$ for all $j \in \{1, \ldots, k\}$, otherwise
$\tre{x} \neq x(\tre{y}_1, \ldots, \tre{y}_k)$. However, that implies that
$\tre{y}_j \in \mathcal{L}(\mathcal{G}\vert B_j)$ for all $j \in \{1, \ldots, k\}$.
The length restriction to $n_j$ follows because---per definition of the tree size---the
sizes $\vert \tre{y}_j\vert $ must add up to $\vert \tre{x}\vert -1$.

Conversely, if there exist nonterminals $B_1, \ldots, B_k \in \Phi$ such that
$r = A \to x(B_1, \ldots, B_k) \in R$ and $\tre{y}_j \in \mathcal{L}_{n_j}(\mathcal{G}\vert B_j)$
for all $j \in \{1, \ldots, k\}$, then there must exist rule sequences
$\bar r_1, \ldots, \bar r_k$ which generate $\tre{y}_j$ from $B_j$ for all
$j \in \{1, \ldots, k\}$. Accordingly, $r, \bar r_1, \ldots, \bar r_k$ generates
$\tre{x}$ from $A$ and, hence, $\tre{x} \in \mathcal{L}(\mathcal{G}\vert A)$.
The length restriction follows because $\vert \tre{x}\vert  = 1 + \vert \tre{y}_1\vert  + \ldots + \vert \tre{y}_k\vert 
\leq 1 + n_1 + \ldots + n_k = n$.
\end{proof}
\end{lma}

\subsection{Deterministic grammars imply unique rule sequences}
\label{ap:unique}

Recall that we define a regular tree grammar as \emph{deterministic} if
no two rules have the same right-hand-side. Our goal is to show
that every deterministic regular tree grammar is unambiguous, in the sense
that there always exists a unique rule sequence generating the tree.
We first prove an auxiliary result.

\begin{lma}
Let $\mathcal{G} = (\Phi, \Sigma, R, S)$ be a deterministic regular tree grammar.
Then, for any two $A \neq A' \in \Phi$, $\mathcal{L}(\mathcal{G}\vert A) \cap
\mathcal{L}(\mathcal{G}\vert A') = \emptyset$.
\begin{proof}
We perform a proof via induction over the tree size. First, consider
trees $\tre x$ with $\lvert \tre x \rvert = 1$, that is, $\tre x = x()$.
If $\tre x \in \mathcal{L}(\mathcal{G}\vert A)$ for some $A \in \Phi$,
the rule $A \to x()$ must be in $R$. Because $\mathcal{G}$ is deterministic,
there can exist no $A' \neq A \in \Phi$ with $A' \to x() \in R$, otherwise
there would be two rules with the same right-hand-side. Accordingly,
$\tre x$ lies in $\mathcal{L}(\mathcal{G}\vert A)$ for at most one $A \in \Phi$.

Now, assume that the claim holds for all trees up to size $n$ and consider
a tree $\tre x$ with $\lvert \tre x \rvert = n+1$. Without loss of generality,
let $\tre x = x(\tre y_1, \ldots, \tre y_k)$. If $\tre x \in \mathcal{L}(\mathcal{G}\vert A)$ for some $A \in \Phi$,
a rule of the form $A \to x(B_1, \ldots, B_k)$ must be in $R$, such that
for all $j \in \{1, \ldots, k\}$:
$\tre y_j \in \mathcal{L}(\mathcal{G}\vert B_j)$.
Now, note that $\lvert \tre y_j \rvert \leq n$. Accordingly, our induction
hypothesis applies and there exists no other $B'_j \neq B_j$ such that
$\tre y_j \in \mathcal{L}(\mathcal{G}\vert B'_j)$. Further, there can exist
no $A' \neq A$ with $A' \to x(B_1, \ldots, B_k)$, otherwise there would be two
rules with the same right-hand-side. Accordingly, $\tre x$  lies in
$\mathcal{L}(\mathcal{G}\vert A)$ for at most one $A \in \Phi$.
\end{proof}
\end{lma}

Now, we prove the desired result.

\begin{thm}\label{thm:unique}
Let $\mathcal{G} = (\Phi, \Sigma, R, S)$ be a deterministic regular tree grammar.
Then, for any $\tre x \in \mathcal{L}(\mathcal{G})$ there exists exactly
one sequence of rules $r_1, \ldots, r_T \in R$ which generates $\tre x$
from $S$.
\begin{proof}
In fact, we will prove a more general result, namely that the claim holds
for all $\tre x \in \bigcup_{A \in \Phi} \mathcal{L}(\mathcal{G}\vert A)$.
We perform an induction over the tree size.

First, consider trees $\tre x$ with $\lvert \tre x \rvert = 1$, that is,
$\tre x = x()$. Then, the only way to generate $\tre x$ is by applying a
single rule of the form $A \to x()$. Because $\mathcal{G}$ is deterministic,
only one such rule can exist. Therefore, the claim holds.

Now,  assume that the claim holds for all trees up to size $n$ and consider
a tree $\tre x$ with $\lvert \tre x \rvert = n+1$. Without loss of generality,
let $\tre x = x(\tre y_1, \ldots, \tre y_k)$. If
$\tre x \in \mathcal{L}(\mathcal{G}\vert A)$ for some $A \in \Phi$, then
there must exist a rule of the form $A \to x(B_1, \ldots, B_k)$ in $R$,
such that for all $j \in \{1, \ldots, k\}$:
$\tre y_j \in \mathcal{L}(\mathcal{G}\vert B_j)$.
Due to the previous lemma and our induction hypothesis we know that for each
$j$, there exists a unique nonterminal $B_j$ and rule sequence $\bar r_j$,
such that $\bar r_j$ generates $\tre y_j$ from $B_j$.
Further, because $\mathcal{G}$ is deterministic, there can exist no other
$A' \neq A \in \Phi$, such that $A' \to x(B_1, \ldots, B_k)$ in $R$.
Therefore, the rule $A \to x(B_1, \ldots, B_k)$ concatenated with the
rule sequences $\bar r_1$, $\ldots$, $\bar r_k$ is the only way to generate
$\tre x$, as claimed.
\end{proof}
\end{thm}

\subsection{All regular tree grammars can be made deterministic}
\label{ap:deterministic}

\begin{thm}
Let $\mathcal{G} = (\Phi, \Sigma, R, \mathcal{S})$ be a regular tree grammar.
Then, there exists a regular tree grammar $\mathcal{G}' = (\Phi', \Sigma, R', \mathcal{S}')$
which is deterministic and where $\mathcal{L}(\mathcal{G}) = \mathcal{L}(\mathcal{G}')$.

\begin{proof}
This proof is an adaptation of Theorem 1.1.9 of \cite{Comon2008} and can also be seen as
an analogue to conversion from nondeterministic finite state machines to deterministic
finite state machines.

In particular, we convert $\mathcal{G}$ to $\mathcal{G}'$ via the following procedure.

We first initialize $\Phi'$ and $R'$ as empty sets.

Second, iterate over $k$, starting at zero up to the maximum number of children in a
rule in $R$, and iterate over all right-hand-sides $x(B_1, \ldots, B_k)$ that occur
in rules in $R$. Next, we collect all $A_1, \ldots, A_m \in \Phi$ such that
$A_i \to x(B_1, \ldots, B_k) \in R$ and add the set $A' = \{A_1, \ldots, A_m\}$ to $\Phi'$.

Third, we perform the same iteration again, but this time we add grammar rules
$\{A_1, \ldots, A_m\} \to x(B_1', \ldots, B_k')$ to $R'$ for all
combinations $B'_1, \ldots, B'_k \in \Phi'$ where $B_j \in B'_j$ for all
$j \in \{1, \ldots, k\}$.

Finally, we define $\mathcal{S}' = \{ A' \in \Phi' \vert A' \cap \mathcal{S} \neq \emptyset\}$.

It is straightforward to see that $\mathcal{G}'$ is deterministic because all rules with
the same right-hand-side get replaced with rules with a unique left-hand-side.

It remains to show that the generated tree languages are the same. To that end, we first
prove an auxiliary claim, namely that for all $A \in \Phi$ it holds:
A tree $\tre{x}$ is in $\mathcal{L}_n(\mathcal{G}\vert A)$ if and only if there exists an
$A' \in \Phi'$ with $A \in A'$ and $\tre{x} \in \mathcal{L}_n(\mathcal{G}'\vert A')$.

We show this claim via induction over $n$. First, consider $n = 1$.
If $\tre{x} \in \mathcal{L}_1(\mathcal{G}\vert A)$, $\tre{x}$ must have the shape $\tre{x} = x()$,
otherwise $\vert \tre{x}\vert  > 1$, and the rule $A \to x()$ must be in $R$. Accordingly,
our procedure assures that there exists some $A' \in \Phi'$ with $A \in A'$ and
$A' \to x() \in R'$. Hence, $\tre{x} \in \mathcal{L}_1(\mathcal{G}'\vert A')$.

Conversely, if $\tre{x} \in \mathcal{L}_1(\mathcal{G}'\vert A')$ for some $A' \in \Phi'$
then $\tre{x}$ must have the shape $\tre{x} = x()$, otherwise $\vert \tre{x}\vert  > 1$, and the
rule $A' \to x()$ must be in $R'$. However, this rule can only be in $R'$ if for any
$A \in A'$ the rule $A \to x()$ was in $R$.
Accordingly, $\tre{x} \in \mathcal{L}_1(\mathcal{G}\vert A)$.

Now, consider the case $n > 1$ and let $\tre{x} = x(\tre{y}_1, \ldots, \tre{y}_k)$.
If $\tre{x} \in \mathcal{L}_n(\mathcal{G}\vert A)$, Lemma~\ref{lma:grammar} tells us that
there must exist nonterminals $B_1, \ldots, B_k \in \Phi$ such that
$A \to x(B_1, \ldots, B_k) \in R$ and $\tre{y}_j \in \mathcal{L}_{n_j}(\mathcal{G}\vert B_j)$
for all $j \in \{1, \ldots, k\}$ with $n_1 + \ldots + n_k = n-1$.
Hence, by induction we know that there exist $B'_j \in \Phi'$ such that $B_j \in B'_j$ and
$\tre{y}_j \in \mathcal{L}_{n_j}(\mathcal{G}'\vert B'_j)$ for all $j \in \{1, \ldots, k\}$.
Further, our procedure for generating $\mathcal{G}'$ ensures that there exists some
$A' \in \Phi'$ such that the rule $A' \to x(B'_1, \ldots, B'_k)$ is in $R'$. Hence,
Lemma~\ref{lma:grammar} tells us that $\tre{x} \in \mathcal{L}_n(\mathcal{G}'\vert A')$.

Conversely, if $\tre{x} \in \mathcal{L}_n(\mathcal{G}'\vert A')$ then Lemma~\ref{lma:grammar}
tells us that there must exist nonterminals $B'_1, \ldots, B'_k \in \Phi'$
such that $A' \to x(B'_1, \ldots, B'_k) \in R'$ and $\tre{y}_j \in \mathcal{L}_{n_j}(\mathcal{G}'\vert B'_j)$
for all $j \in \{1, \ldots, k\}$ with $n_1 + \ldots + n_k = n-1$.
Hence, by induction we know that for all $j \in \{1, \ldots, k\}$ and all $B_j \in B'_j$
we obtain $\tre{y}_j \in \mathcal{L}_{n_j}(\mathcal{G}\vert B_j)$.
Further, our procedure for generating $\mathcal{G}'$ ensures that for any $A \in A'$
there must exist a rule $A \to x(B_1, \ldots, B_k) \in R$ for some $B_1, \ldots, B_k
\in \Phi$ and $B_j \in B'_j$ for some combination of $B'_j \in \Phi'$. Hence,
Lemma~\ref{lma:grammar} tells us that $\tre{x} \in \mathcal{L}_n(\mathcal{G}\vert A)$.
This concludes the induction.

Now, note that $\tre{x} \in \mathcal{L}(\mathcal{G})$ implies that there exists
some $S \in \mathcal{S}$ with $\tre{x} \in \mathcal{L}(\mathcal{G}\vert S)$. Our auxiliary
result tells us that there then exists some $S' \in \Phi'$ with $S \in \Phi'$ and
$\tre{x} \in \mathcal{L}(\mathcal{G}'\vert S')$. Further, since $S' \cup \mathcal{S}$ contains
at least $S$, our construction of $\mathcal{G}'$ implies that $S' \in \mathcal{S}'$.
Hence, $\tre{x} \in \mathcal{L}(\mathcal{G}')$.

Conversely, if $\tre{x} \in \mathcal{L}(\mathcal{G}')$, there must exist some
$S' \in \mathcal{S}'$ with $\tre{x} \in \mathcal{L}(\mathcal{G}'\vert S')$.
Since $S' \in \mathcal{S}'$, there must exist at least one $S \in S'$ such that
$S \in \mathcal{S}$. Further, our auxiliary result tells us that $\tre{x} \in \mathcal{L}(\mathcal{G}\vert S)$. Accordingly, $\tre{x} \in \mathcal{L}(\mathcal{G})$.
\end{proof}
\end{thm}

As an example, consider the following grammar $\mathcal{G} = (\Phi, \Sigma, R, \mathcal{S})$
for logical formulae in conjuctive normal form.
\begin{align*}
\Phi = \{&F, C, L, A\} \\
\Sigma = \{&\wedge, \vee, \neg, x, y\} \\
R = \{&F \to \wedge(C, F) \vert \vee(L, C) \vert \neg(A) \vert x \vert y, \\
& C \to \vee(L, C) \vert \neg(A) \vert x \vert y, \\
& L \to \neg(A) \vert x \vert y, \\
& A \to x \vert y\} \\
\mathcal{S} = \{&F\}
\end{align*}

The deterministic version of this grammar is $\mathcal{G}' = (\Phi', \Sigma, R', \mathcal{S}')$
with $\mathcal{S}' = \Phi'$ and
\begin{align*}
\Phi' = \{&\{F, C, L, A\}, \{F, C, L\}, \{F, C\}, \{F\}\} \\
R' = \Big\{&\{F, C, L, A\} \to x \vert y, \\
&\{F, C, L\} \to \neg(\{F, C, L, A\}), \\
&\{F, C\} \to \vee(\{F, C, L, A\}, \{F, C, L, A\}), \\
&\{F, C\} \to \vee(\{F, C, L, A\}, \{F, C, L\}), \\
&\{F, C\} \to \vee(\{F, C, L, A\}, \{F, C\}), \\
&\{F, C\} \to \vee(\{F, C, L\}, \{F, C, L, A\}), \\
&\{F, C\} \to \vee(\{F, C, L\}, \{F, C, L\}), \\
&\{F, C\} \to \vee(\{F, C, L\}, \{F, C\}), \\
&\{F\} \to \wedge(\{F, C, L, A\}, \{F, C, L, A\}), \\
&\{F\} \to \wedge(\{F, C, L, A\}, \{F, C, L\}), \\
&\{F\} \to \wedge(\{F, C, L, A\}, \{F, C\}), \\
&\{F\} \to \wedge(\{F, C, L, A\}, \{F\}), \\
&\{F\} \to \wedge(\{F, C, L\}, \{F, C, L, A\}), \\
&\{F\} \to \wedge(\{F, C, L\}, \{F, C, L\}), \\
&\{F\} \to \wedge(\{F, C, L\}, \{F, C\}), \\
&\{F\} \to \wedge(\{F, C, L\}, \{F\}), \\
&\{F\} \to \wedge(\{F, C\}, \{F, C, L, A\}), \\
&\{F\} \to \wedge(\{F, C\}, \{F, C, L\}), \\
&\{F\} \to \wedge(\{F, C\}, \{F, C\}), \\
&\{F\} \to \wedge(\{F, C\}, \{F\})
\Big\}.
\end{align*}

\subsection{Proof of the Parsing Theorem}
\label{ap:parsing}

We first generalize the statement of the theorem to also work for regular tree grammars
with multiple starting symbols. In particular, we show the following:

\begin{thm}
Let $\mathcal{G} = (\Phi, \Sigma, R, \mathcal{S})$ be a deterministic regular tree grammar.
Then, it holds: $\tre x$ is a tree in $\mathcal{L}(\mathcal{G})$ if and only if
Algorithm~\ref{alg:encoding} returns a nonterminal $S \in \mathcal{S}$, a unique rule
sequence $r_1, \ldots, r_T$ that generates $\tre x$, and a vector $\Enc(\tre x) \in \R^n$
for the input $\tre x$.
Further, Algorithm~\ref{alg:encoding} has $\mathcal{O}(\vert \tre x\vert )$ time and space complexity.
\begin{proof}
We structure the proof in three parts. First, we show that $\tre x \in \mathcal{L}(\mathcal{G})$
implies that Algorithm~\ref{alg:encoding} returns with a nonterminal $S \in \mathcal{S}$, a unique
generating rule sequence, and some vector $\Enc(\vec x) \in \R^n$. Second, we show
that if Algorithm~\ref{alg:encoding} returns with a nonterminal $S \in \mathcal{S}$ and some rule
sequence as well as some vector, then the tree generated by the rule sequence lies in
$\mathcal{L}(\mathcal{G})$. Finally, we prove the complexity claim.

We begin by a generalization of the first claim.
For all $A \in \Phi$ and for all $\tre x \in \mathcal{L}(\mathcal{G}\vert A)$ it holds:
Algorithm~\ref{alg:encoding} returns the nonterminal $A$, a rule sequence
$r_1, \ldots, r_T \in \R$ that generates $\tre x$ from $A$, and a vector $\Enc(\tre x)$.
The uniqueness of the rule sequence $r_1, \ldots, r_T$ follows from
Theorem~\ref{thm:unique}.

Our proof works via induction over the tree size.
Let $A \in \phi$ be any nonterminal and let $\tre x \in \mathcal{L}_1(\mathcal{G}\vert A)$.
Then, because $\vert \tre x\vert  = 1$, $\tre x$ must have the form $\tre x = x()$ for some $x \in \Sigma$
and there must exist a (unique) rule of the form $r = A \to x() \in R$ for some $A \in \Phi$, otherwise
$\tre x$ would not be in $\mathcal{L}_1(\mathcal{G}\vert A)$. 
Now, inspect Algorithm~\ref{alg:encoding} for $\tre x$. Because $k = 0$, lines 2-4 do not get
executed. Further, because $\mathcal{G}$ is deterministic,
the $\exists$ Quantifier in line 5 is unique, such that $r$ in line 6
is exactly $A \to x()$. Accordingly, line 7 returns exactly $A, r, \enc^r$.

Now, assume that the claim holds for all trees $\tre y \in \bigcup_{A \in \Phi}
\mathcal{L}_n(\mathcal{G}\vert A)$ with $n \geq 1$. Then, consider some nonterminal $A \in \Phi$
and some tree $\tre x \in \mathcal{L}_{n+1}(\mathcal{G}\vert A)$ with size $\vert \tre x\vert  = n + 1$.
Because $\vert \tre x\vert  > 1$, $\tre x$ must have the shape $\tre x = x(\tre y_1, \ldots, \tre y_k)$
with $k > 0$ for some $x \in \Sigma$ and some trees $\tre y_1, \ldots, \tre y_k$.
Lemma~\ref{lma:grammar} now tells us that there must exist nonterminals $B_1, \ldots, B_k
\in \Phi$ with $\tre y_j \in \mathcal{L}_{n_j}(\mathcal{G}\vert B_j)$ for all $j \in \{1, \ldots, k\}$
such that $n_1 + \ldots + n_k = n-1$. Hence, our induction applies to all trees $\tre y_1, \ldots,
\tre y_k$.

Now, inspect Algorithm~\ref{alg:encoding} once more. Due to
induction we know that line 3 returns for each tree $\tre y_j$ a unique combination of
nonterminal $B_j$ and rule sequence $\bar r_j$, such that $\bar r_j$ generates $\tre y_j$ from
$B_j$. Further, the rule $A \to x(B_1, \ldots, B_k)$ must exist in $R$, otherwise $\tre x \notin
\mathcal{L}_{n+1}(\mathcal{G}\vert A)$. Also, there can not exist another nonterminal $B \in \Phi$
with $B \to x(B_1, \ldots, B_k) \in R$, otherwise $\mathcal{G}$ would not be deterministic.
Therefore, $r$ in line 6 is exactly $A \to x(B_1, \ldots, B_k)$. It remains to show that
the returned rule sequence does generate $\tre x$ from $A$. The first step in our generation is
to pop $A$ from the stack, replace $A$ with $x(B_1, \ldots, B_k)$, and push $B_k, \ldots, B_1$
onto the stack. Next, $B_1$ will be popped from the stack and the next rules will be $\bar r_1$.
Due to induction we know that $\bar r_1$ generates $\tre y_1$ from $B_1$, resulting in the
intermediate tree $x(\tre y_1, B_2, \ldots, B_k)$ and the stack $B_k, \ldots, B_2$. Repeating this
argument with the remaining rule sequence $\bar r_2, \ldots, \bar r_k$ yields exactly $\tre x$ and
an empty stack, which means that the rule sequence does indeed generate $\tre x$ from $A$.

For the second part, we also consider a generalized claim. In particular, if
Algorithm~\ref{alg:encoding} for some input tree $\tre x$ returns with some nonterminal $A$ and
some rule sequence $\bar r$, then $\bar r$ generates $\tre x$ from $A$.
We again prove this via induction over the length of the input tree. If $\vert \tre x\vert  = 1$, then
$\tre x = x()$ for some $x$. Because $k = 0$, lines 2-4 of Algorithm~\ref{alg:encoding} do not
get executed. Further, there must exist some nonterminal $A \in \Phi$ such that
$r = A \to x() \in R$, otherwise Algorithm~\ref{alg:encoding} would
return an error, which is a contradiction. Finally, the return values are $A$ and $r$ and
and $r$ generates $\tre x = x()$ from $A$ as claimed.

Now, assume the claim holds for all trees with length up to $n \geq 1$ and consider a tree
$\tre x = x(\tre y_1, \ldots, \tre y_k)$ with $\vert \tre x\vert  = n + 1$. Because $n+1 > 1$, $k > 0$.
Further, for all $j \in \{1, \ldots, k\}$, $\vert \tre y_j\vert  < \vert \tre x\vert  = n + 1$, such that the
induction hypothesis applies. Accordingly, line 3 returns nonterminals $B_j$ and rule sequences
$\bar r_j$, such that $\bar r_j$ generates $\tre y_j$ from $B_j$ for all $j \in \{1, \ldots, k\}$.
Further, some nonterminal $A \in \Phi$ must exist such that $A \to x(B_1, \ldots, B_k) \in R$,
otherwise Algorithm~\ref{alg:encoding} would not return, which is a contradiction. Finally,
the return values are $A$ and $r, \bar r_1, \ldots, \bar r_k$ and $r, \bar r_1, \ldots, \bar r_k$
generates $\tre x$ from $A$ using the same argument as above. This concludes the proof by induction.

Our actual claim now follows. In particular, if $\tre x \in \mathcal{L}(\mathcal{G})$,
then there exists some $S \in \mathcal{S}$ such that $\tre x \in \mathcal{L}(\mathcal{G}\vert S)$.
Accordingly, Algorithm~\ref{alg:encoding} returns with $S$,
a generating rule sequence from $S$, and a vector $\Enc(\tre x)$ as claimed.
Conversely, if for some input tree $\tre x$ Algorithm~\ref{alg:encoding} returns
with some nonterminal $S \in \mathcal{S}$ and rule sequence $\bar r$, then $\bar r$
generates $\tre x$ from $S$, which implies that $\tre x \in \mathcal{L}(\mathcal{G})$.

Regarding time and space complexity, notice that each rule adds exactly one node tot he tree and that each recursion of Algorithm~\ref{alg:encoding} adds exactly one rule. Accordingly,
Algorithm~\ref{alg:encoding} requires exactly $\vert \tre x\vert $ iterations. Since the returned rule sequence
has length $\vert \tre x\vert $ and all other variables have constant size, the space complexity
is the same.
\end{proof}
\end{thm}

\section{Determinism for Optional and Starred Nonterminals}

As mentioned in the main text, $A?$ denotes a nonterminal with the
production rules $A? \to A$ and $A? \to \varepsilon$, and $A*$ denotes a nonterminal
with the production rules $A* \to A, A*$ and $A* \to \varepsilon$, where the comma
refers to concatenation. It is easy to see that these concepts can make a grammar
ambiguous. For example, the grammar rule $A \to x(B*, B*)$ is ambiguous because for
the partially parsed tree $x(B)$ we could not decide whether $B$ has been produced by the
first or second $B*$. Similarly, the two grammar rules $A \to x(B*, C)$ and $B \to x(C)$
are ambiguous because the tree $x(C)$ can be produced by both.

More generally, we define a regular tree grammar as deterministic if the following two
conditions are fulfilled.
\begin{enumerate}
\item Any grammar rule $A \to x(B_1, \ldots, B_k) \in R$ must be internally deterministic
in the sense that any two neighboring symbols $B_j$, $B_{j+1}$ must be either unequal or
not both starred/optional.
\item Any two grammar rules $A \to x(B_1, \ldots, B_k) \in R$ and
$A' \to x(B'_1, \ldots, B'_l) \in R$ must be non-intersecting, in the sense that the
sequences $B_1, \ldots, B_k$ and $B'_1, \ldots, B'_l$, when interpreted as regular
expressions, are not allowed to have intersecting languages.
\end{enumerate}

Note that, if we do not have starred or optional nonterminals in any rule, the first
condition is automatically fulfilled and the second condition collapses to our original
definition of determinism, namely that no two rules are permitted to have the same
right hand side. Further note that the intersection of regular expression languages
can be checked efficiently via standard techniques from theoretical computer science.

If both conditions are fulfilled, we can adapt Algorithm~\ref{alg:encoding} in a straightforward
fashion by replacing line 5 with $\exists A \in \Phi, A \to x(B'_1, \ldots, B'_l) \in R$
such that $B'_1, \ldots, B'_l$ when interpreted as a regular expression matches $B_1, \ldots, B_k$.
There can be only one such rule, otherwise the second condition would not be fulfilled.
We then the encodings $\vec y_1, \ldots, \vec y_k$ to their matching nonterminals in the regular
expression $B'_1, \ldots, B'_l$ and proceed as before.

\end{appendix}

\end{document}